\documentclass[10pt, conference, compsocconf]{IEEEtran}
\usepackage{booktabs}
\usepackage{url}
\usepackage{graphicx}
\usepackage{algorithm}
\usepackage{multirow}
\usepackage[noend]{algpseudocode}
\usepackage{hyperref}
\usepackage[utf8]{inputenc}
\makeatletter
\def\BState{\State\hskip-\ALG@thistlm}
\makeatother
\title{\textit{Expect the unexpected}: Harnessing Sentence Completion for Sarcasm Detection}
\author{\begin{tabular}{cccc}
\textbf{Aditya Joshi$^{1,2,3}$} & \textbf{Samarth Agrawal$^{1}$} & \textbf{Pushpak Bhattacharyya$^{1}$} & \textbf{Mark Carman$^{2}$}\\
\end{tabular}\\
\begin{tabular}{cc}
\multicolumn{1}{c}{$^{1}$Indian Institute of Technology Bombay, India} & \multicolumn{1}{c}{$^{2}$Monash University, Australia}\\
\multicolumn{2}{c}{$^{3}$IITB-Monash Research Academy, India}\\
\multicolumn{2}{c}{\tt \{adityaj,samartha,pb\}@cse.iitb.ac.in, mark.carman@monash.edu}
\end{tabular}
}
\begin{document}
\maketitle
\begin{abstract}
The trigram `\textit{I love being}' is expected to be followed by positive words such as `\textit{happy}'. In a sarcastic sentence, however, the word `\textit{ignored}' may be observed. The expected and the observed words are, thus, incongruous. We model sarcasm detection as the task of detecting incongruity between an observed and an expected word. In order to obtain the expected word, we use Context2Vec, a sentence completion library based on Bidirectional LSTM. However, since the exact word where such an incongruity occurs may not be known in advance, we present two approaches: an All-words approach (which consults sentence completion for every content word) and an Incongruous words-only approach (which consults sentence completion for the 50\% most incongruous content words). The approaches outperform reported values for tweets but not for discussion forum posts. This is likely to be because of redundant consultation of sentence completion for discussion forum posts. Therefore, we consider an oracle case where the exact incongruous word is manually labeled in a corpus reported in past work. In this case, the performance is higher than the all-words approach. This sets up the promise for using sentence completion for sarcasm detection.
\end{abstract}
This paper has been selected for publication at 2017 CONFERENCE OF THE PACIFIC ASSOCIATION FOR COMPUTATIONAL LINGUISTICS (\url{http://pacling.ucsy.edu.mm/pacling/}) 
\section{Introduction}
Sarcasm is defined as ``\textit{the use of irony to mock or convey contempt}\footnote{Source: Oxford Dictionary}". For example, the sentence `\textit{I love being ignored}' is sarcastic. Automatic sarcasm detection is the task of predicting whether or not a given text contains sarcasm. Several statistical approaches have been proposed for sarcasm detection~\cite{3}~\cite{13}~\cite{47}. In addition, rule-based approaches based on evidences of sarcasm have also done well~\cite{khattri}~\cite{similes}~\cite{maynard}. This paper presents another rule-based technique. Our technique is novel in its application of sentence completion for sarcasm detection. 

As an introduction to the technique, consider the sarcastic sentence `\textit{I love being ignored}'. A likely word to follow the trigram `\textit{I love being}' would be a positive sentiment word such as `\textit{happy}'. However, in the sarcastic sentence, the word `\textit{ignored}' occurs. The word `\textit{ignored}' in the sarcastic sentence is semantically distant from an expected word such as `\textit{happy}'. This (dis)similarity can be used as an indicator of incongruity which is central to sarcasm, as per linguistic studies~\cite{gibbs1994}\cite{ivanko2003context}. In order to obtain the expected word at a given position, we harness automatic sentence completion. Sentence completion predicts the most likely word at a given position in a sentence~\cite{zweig2011microsoft}. For our experiments, we use  context2vec, a sentence completion toolkit~\cite{context2vec}. Thus, our paper deals with the question:
\\\\
\textit{Because incongruity in sarcasm is a phenomenon where the unexpected is observed, can sarcasm be detected using sentence completion?}
\\\\
A key assumption here is that a sentence completion toolkit trained on a large, general-purpose corpus follows the language model for non-sarcastic text. The assumption is reasonable because the sentence completion model is likely to have learned the language model for non-sarcastic text since sarcasm is an infrequent phenomenon. 

It must be noted that the exact observed word where the incongruity occurs (`\textit{ignored}' in the example above) is not known in advance. Hence, a sentence contains multiple candidate words of incongruity, out of which the incongruity is observed in case of specific word(s). We refer to these words as the `\textit{incongruous word(s)}'. Therefore, our approaches vary in terms of the candidate incongruous words that are considered. 

The novelty of this paper is as follows:
\begin{enumerate}
\item Using sentence completion for sarcasm detection
\item Experimentation with short text (where candidate incongruous words are a small set of words), long text (where candidate incongruous words are a large set of words), and an oracle case (where the exact incongruous word is known)
\end{enumerate}
The rest of the paper is organized as follows. Section~\ref{sec:relwork} describes the related work. Section~\ref{sec:motiv} presents the motivation behind using sentence completion. Section~\ref{sec:appr} presents two approaches: an all-words approach, and an incongruous words-only approach. As stated earlier, the two approaches differ in terms of candidate incongruous words. Section~\ref{sec:expsetup} gives the experiment setup while Section~\ref{sec:res} presents the results. We discuss an oracle case scenario in Section~\ref{sec:disc} to validate the strength of our hypothesis. Finally, we analyze the errors made by our system in Section~\ref{sec:erranal} and conclude the paper in Section~\ref{sec:concl}.
\section{Related Work}
\label{sec:relwork}
Majority of the past work in statistical sarcasm detection uses sarcasm-specific features such as punctuations, emoticons or sarcasm-indicating n-grams ~\cite{3}\cite{13}\cite{joshi2015harnessing}\cite{riloff}\cite{47}. For example, ~\cite{3} present a semi-supervised algorithm that first extracts sarcasm-indicating n-grams and then use them as features for a classifier. ~\cite{joshi2015harnessing} use features based on number of sentiment flips, positive/negative subsequences, in addition to such n-grams. ~\cite{47} include features such as audience information, twitter familiarity, etc. 

Recent work in sarcasm detection employs features that capture contextual information such as an author's background, conversational context, etc.~\cite{24}\cite{26}\cite{29}\cite{41}\cite{43}. Formulations beyond classifiers have also been considered. For example, \cite{43} use sequence labeling algorithms to predict sarcasm in individual utterances in a dialogue. On the other hand, ~\cite{41} use them to predict sarcasm of the last utterance in a dialogue with automatic labels in the rest of the sequence. However, in our case, we do not use any contextual information from the author or the conversation. This means that a hyperbolic sentence such as `\textit{X is the best President ever!}' (where the sarcasm cannot be understood based on the text alone) is beyond the scope of our approach. 

In addition to the above, several rule-based techniques based on intuitive indicators of sarcasm have been reported. \cite{maynard} predict a tweet as sarcastic if sentiment in the text of the tweet contradicts with the sentiment of a hashtag in the tweet. \cite{khattri} predict a tweet as sarcastic if sentiment of the tweet does not match with sentiment of past tweets by the author of the tweet towards the entities in the tweet. Similarly, \cite{similes} use a set of nine rules to predict if a given simile (for example, `\textit{as exciting as a funeral}') is sarcastic.~\cite{riloff} capture sarcasm as a combination of positive verbs followed by negative situation phrases. Our approach is rule-based as well.

Our work is the first to employ sentence completion for the purpose of sarcasm detection. Sentence completion approaches based on word embeddings have been reported~\cite{mikolov2013efficient}\cite{liu2015learning}. However, they are only for sentence completion and not for sarcasm detection. They restrict themselves to completing sentences. We propose and validate the hypothesis that a `language model incongruity' as experienced by a sentence completion module can be useful for sarcasm detection.  We use context2vec~\cite{context2vec} as the sentence completion library. The distinction between these sentence completion approaches is beyond the scope of this paper because the focus is to use one of them for sarcasm detection and demonstrate that it works.
\section{Motivation}
\label{sec:motiv}
As stated in the previous section, in the sarcastic example `\textit{I love being ignored}', the word `\textit{ignored}' is observed at a position where positive sentiment words would be expected. Hence, the word `\textit{ignored}' is the exact incongruous word. Specifically, if context2vec~\cite{context2vec} were consulted to complete the sentence `\textit{I love being []}' where [] indicates the position for which the most likely word is to be computed, the word `\textit{happy}' is returned. Word2vec similarity between `\textit{happy}' and `\textit{ignored}' is 0.0204, for certain pre-trained word2vec embeddings. This low value of similarity between the expected and observed words can be harnessed as an indicator for sarcasm.  In the rest of the paper, we refer to the word present at a given position as the `\textbf{observed word}' (`\textit{ignored}' in the example above) where the most likely word at the position as returned by sentence completion is the `\textbf{expected word}' (`\textit{happy}' in the example above). 

However, a caveat lies in determination of the candidate incongruous words for which sentence completion will be consulted. For example, the sentence `\textit{I could not make it big in Hollywood because my writing was not bad enough}' is sarcastic because of the incongruous word `\textit{bad}' which is at the penultimate position in the sentence. In the absence of the knowledge of this exact incongruous word, it is obvious that an algorithm must iterate over a set of candidate incongruous words. Hence, we present two approaches: one which iterates over all words and another which restricts to a subset of words. The first approach is called the all-words approach, while the second is incongruous words-only approach. These approaches are described in detail in the next section. The `\textit{oracle case}' for our algorithm is a situation where the incongruous word is exactly known. We validate that our algorithm holds benefit even for the oracle case, in Section~\ref{sec:disc}.


\section{Approach}
\label{sec:appr}
We present two approaches that use sentence completion for sarcasm detection: (a) an ``all-words'' approach, and (b) ``incongruous words-only'' approach. As stated earlier, in the absence of the knowledge about the exact position of incongruity, our technique must iterate over multiple candidate positions. For both the approaches, the following holds:\\ \\
\vspace{.5em}
\noindent\fbox{ \parbox{.99\columnwidth}{
\textbf{Input}: A text of length $l$\\
\textbf{Output}: Sarcastic/non-sarcastic \\
\textbf{Parameters}: 
\begin{itemize} \setlength\itemsep{0cm}
\item Similarity measure $sim(w_i,w_k)$ returning the similarity between words $w_i$ and $w_k$
\item Threshold $T$ (a real value between minimum and maximum value of $sim(w_i,w_k)$)
\end{itemize}
}}
\subsection{All-words approach}
As the name suggests, this approach considers all content words\footnote{Content words are words that are not function words. We ignore function words in a sentence.} as candidate incongruous words. This approach is as follows:

\vspace{.5em}
\noindent\fbox{ \parbox{.97\columnwidth}{ \parindent=.5cm 
\noindent
$min\leftarrow \infty$ \\ 
for $p$ = 1 to $l$ do:\\
\indent\indent \% compute expected word:\\ 
\indent $e_p\leftarrow context2vec(w_1,...,w_{p-1},[],w_{p+1},...,w_l)$ \\
\indent\indent \% check similarity to observed word: \\
\indent if $sim(e_p,w_p)<min$ then $min \leftarrow sim(e_p,w_p)$\\
if $min<T$ then predict \emph{sarcastic}
}}
\vspace{.5em}

Thus, for the sentence `\textit{A woman needs a man like a fish needs a bicycle}\footnote{http://www.phrases.org.uk/meanings/414150.html}' containing five content words (out of which `\textit{needs}' occurs twice), the sentence completion library will be consulted as follows:
\begin{enumerate}\setlength\itemsep{0cm}
\item A [] needs a man like a fish needs a bicycle.
\item A woman [] a man like a fish needs a bicycle.
\item A woman needs a [] like a fish needs a bicycle.
\item A woman needs a man like a [] needs a bicycle.
\item A woman needs a man like a fish [] a bicycle.
\item A woman needs a man like a fish needs a [].
\end{enumerate}

\subsection{Incongruous words-only Approach}
A key shortcoming of the previous approach is that it may use similarity values for words which are not incongruous, since it makes six calls in case of the example given. For example, the first part of the sentence does not contain a language model incongruity and hence, the calls are redundant.   Our second approach, the Incongruous words-only approach, reduces the set of words to be checked by sentence completion to half, thereby eliminating redundant comparisons as shown in the previous subsection. Incongruous words-only approach is as follows:

\vspace{.5em}
\noindent\fbox{ \parbox{.97\columnwidth}{\parindent=.5cm
\noindent
for $p$ = 1 to $l$ do:\\
\indent\indent \% compute average similarity to words:\\ 
\indent $\bar{s}_p\leftarrow \frac{1}{l-1}\sum_{i\neq p}sim(w_i,w_p)$ \\
\indent \% choose positions with lowest averages: \\ 
$Incongruous\leftarrow \{i: \bar{s}_i\leq median(\bar{s}_1,...,\bar{s}_l)\}$ \\
$min\leftarrow \infty$ \\ 
for $p\in Incongruous$ do:\\
\indent\indent \% compute expected word:\\ 
\indent $e_p\leftarrow context2vec(w_1,...,w_{p-1},[],w_{p+1},...,w_l)$ \\
\indent\indent \% check similarity to observed word: \\
\indent if $sim(e_p,w_p)<min$ then $min \leftarrow sim(e_p,w_p)$\\
if $min<T$ then predict \emph{sarcastic}
}}
\vspace{.5em}

As seen above, we first select the required subset of words in the sentence. Beyond that, the approach is the same as the all-words approach. As a result, for the sentence `\textit{A woman needs a man like a fish needs a bicycle}', `\textit{fish}', `\textit{needs}' and `\textit{bicycle}' are returned as most incongruous Incongruous words-only. Hence, the sentence completion is now consulted for the following input strings:

\begin{enumerate}\setlength\itemsep{0cm}
\item A woman [] a man like a fish needs a bicycle.
\item A woman needs a man like a [] needs a bicycle.
\item A woman needs a man like a fish [] a bicycle.
\item A woman needs a man like a fish needs a [].
\end{enumerate}
We hope that this reduction in the set of candidate strings increases the chances of the algorithm detecting the incongruous word and hence, the sarcasm. We observe an interesting trend in short versus long text in terms of this reduction, as will be discussed in the forthcoming sections.
\section{Experiment Setup}
\label{sec:expsetup}
Since our approaches are contingent on the set of candidate phrases being considered, we consider two scenarios: short text where the set of words where incongruity has likely occurred is small, and long text where the set is large. Therefore, the two datasets used for the evaluation of our approaches are: (a) Tweets by~\cite{riloff} (2278 total, 506 sarcastic, manually annotated), and (b) Discussion forum posts by~\cite{iaccorpus} (752 sarcastic, 752 non-sarcastic, manually annotated). We ignore function words when we iterate over word positions. They are not removed because such removal would disrupt the sentence, which is undesirable since we use sentence completion. We use a list of function words available online\footnote{http://www.ranks.nl/stopwords}. 

For both approaches, we repeat the experiments over a range of threshold values, and report the best results (and the corresponding threshold values). As similarity measures, we use (a) \textbf{word2vec similarity} computed using pre-trained embeddings given by the Word2Vec tool. These embeddings were learned on the Google News corpus\footnote{\url{https://code.google.com/archive/p/Word2Vec/}}, (b) \textbf{WordNet similarity} from WordNet::similarity by \cite{pedersen2004WordNet} (specifically, Wu-Palmer Similarity). The word2vec similarity in Incongruous words-only approach is computed in the same manner as word2vec similarity above. Since word2vec similarity may not be low for antonyms, we set the similarity measure for antonyms as 0. As stated earlier, for sentence completion, we use context2vec by ~\cite{context2vec}. It is a sentence completion toolkit that uses Bidirectional LSTM to predict a missing word, given a sentence. We use the top word returned by context2vec, as per the model trained on UkWac corpus\footnote{ \url{http://u.cs.biu.ac.il/~nlp/resources/downloads/context2vec/}}.

We report our evaluation for two configurations:
\begin{enumerate}
\item \textit{Overall Performance}: In the first case, we run the algorithm for a range of threshold values and report results for the complete dataset.
\item \textit{Two-fold cross-validation}: Our algorithm is dependent on the value of the threshold. Hence, we divide the dataset into two splits and repeat the experiments in two runs: estimate the optimal threshold on a split, and report results for the other, and vice versa. 
\end{enumerate}
   \begin{figure}
\centering
        \includegraphics[width=0.46\textwidth]{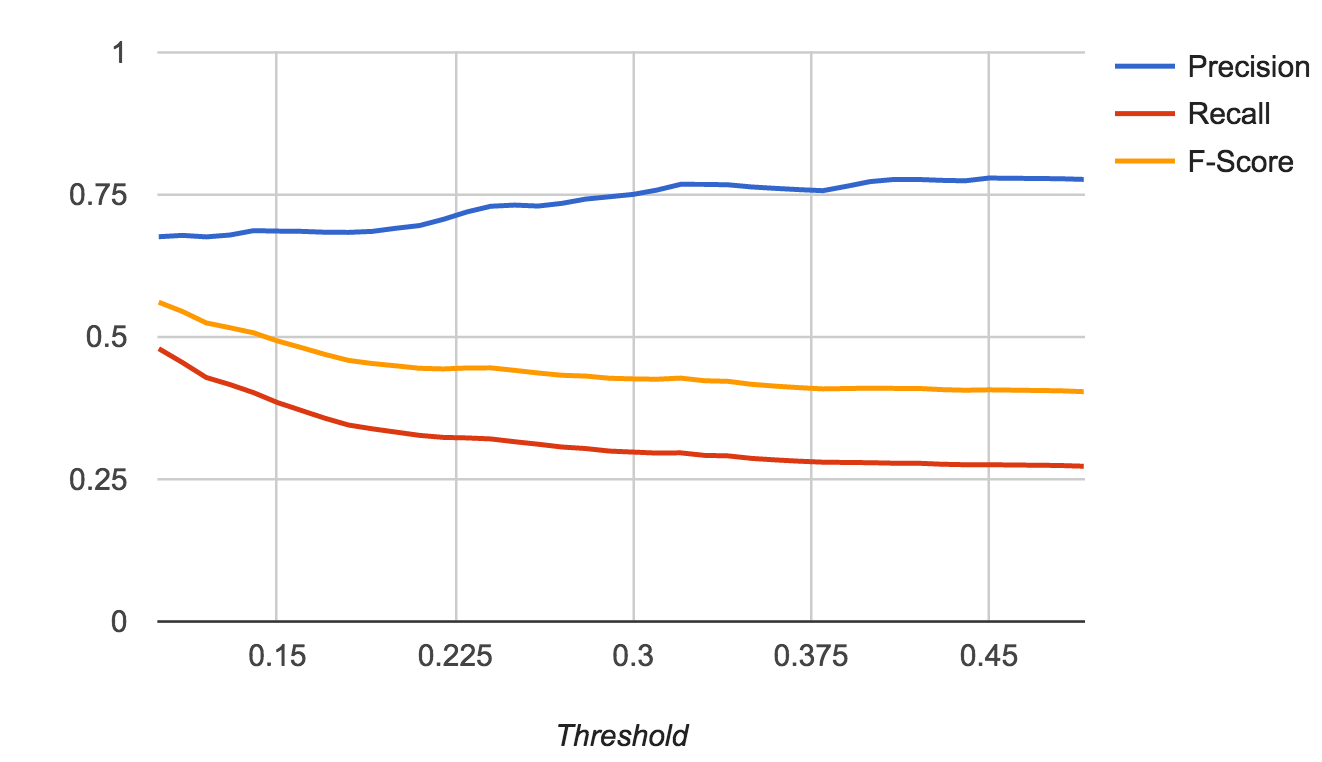}
   \caption{Determining optimal value of threshold; Tweets, word2vec, All-words approach}
   \label{fig:optimal2}
   \end{figure}
\begin{figure}
\centering
        \includegraphics[width=0.46\textwidth]{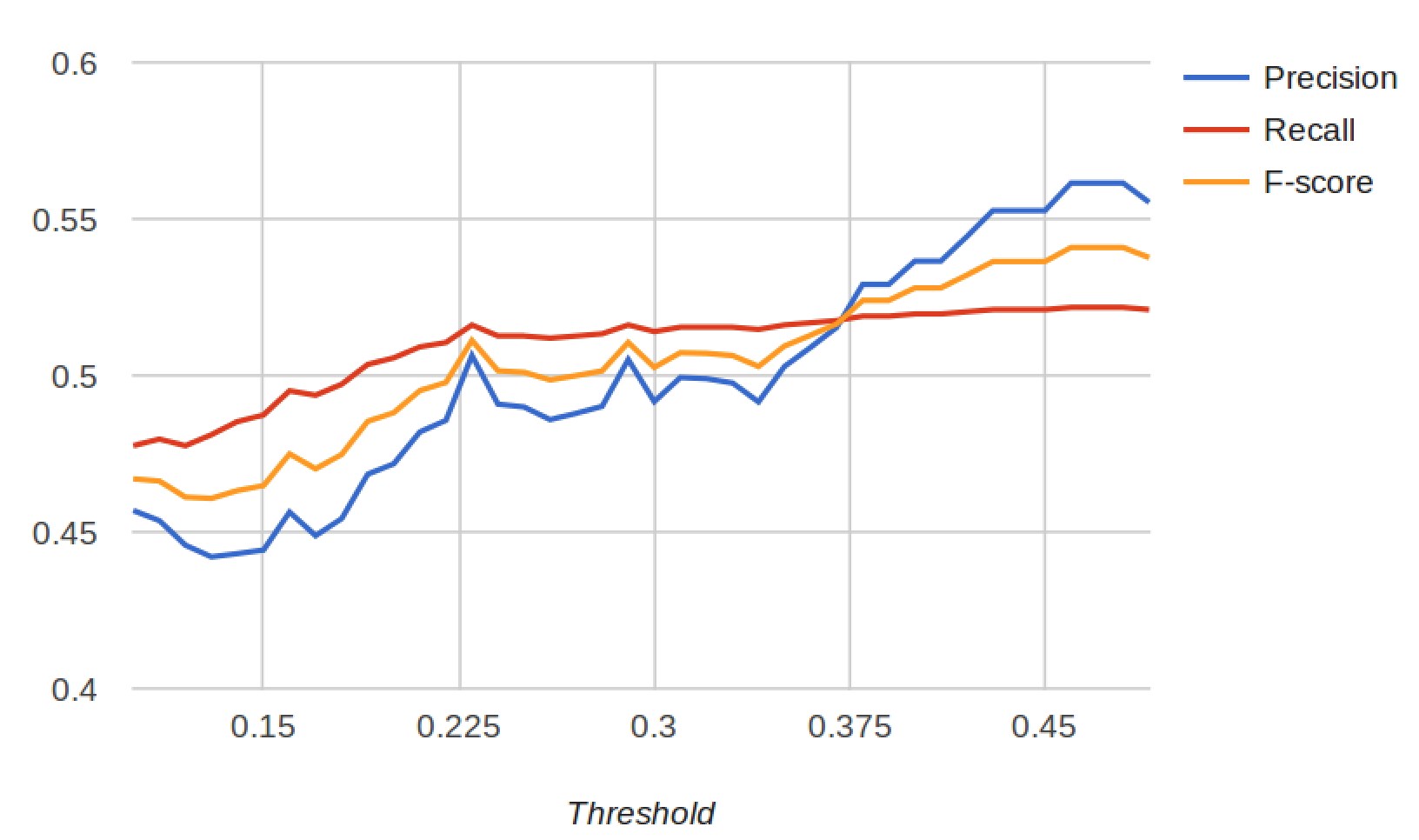}
   \caption{Determining optimal value of threshold; Discussion Forum posts, word2vec, All-words approach}
   \label{fig:optimal}
   \end{figure}
\section{Results}
\label{sec:res}
In this section, we present an evaluation of our approaches, on the two datasets: the first consisting of short text (tweets), and the second consisting of long text (discussion forum posts). We first show the results for the complete dataset with optimal values of the threshold. We then repeat our experiments where the threshold is determined using a train-test split. These two configuations (overall performance and two-fold cross-validation) collectively validate the benefit of our approach.
\subsection{Overall Performance}
Table ~\ref{tab:res1} shows the performance for tweets. Figure~\ref{fig:optimal2} shows how optimal thresholds are determined. When word2vec similarity is used for the all-words approach, an F-score of 54.48\% is obtained. We outperform two past works \cite{riloff,joshi2015harnessing} which have reported their values on the same dataset. The best F-score of 80.24\% is obtained when WordNet is used as the similarity measure in our Incongruous words-only approach. We observe that the Incongruous words-only approach performs significantly better than the all-words approach. In case of word2vec similarity, the F-score increases by around 18\%, and by around 9\% in case of WordNet similarity. Also, the optimal threshold values are lower for the all-words approach as compared to the Incongruous words-only approach.   

\begin{table}[]
\centering

\begin{tabular}{p{2.8cm}p{0.5cm}lll}
\toprule
 	&&	\textbf{P} &	\textbf{R} &	\textbf{F}\\ \midrule
Riloff et al. (2013) & & 62 & 44 & 51 \\ 
Joshi et al. (2015) & & 77 & 51 & 61 \\
    \midrule
   \textbf{Similarity} 	&\textbf{T}&	\textbf{P}&	\textbf{R}&\textbf{F}\\ \midrule
\multicolumn{5}{c}{\textbf{All-Words Approach}}\\\midrule
Word2Vec&	0.11&	67.85&	45.51	&54.48\\
	WordNet&	0.11&	67.68	&74.84	&71.08\\ \midrule
\multicolumn{5}{c}{\textbf{Incongruous words-only Approach}}\\\midrule
Word2Vec	&0.42	&68.00&	77.64&	72.50	\\
	WordNet&	0.12&	82.77&	77.87	&80.24	\\
\bottomrule
\end{tabular}
\caption{Results of our approach for dataset of tweets by Riloff et al. (2013), compared with best reported values (Joshi et al. (2015) and Riloff et al. (2013)) on the same dataset}
\label{tab:res1}
\end{table}


Table~\ref{tab:res2} shows the performance of our approaches for discussion forum posts, compared with past work by ~\cite{joshi2015harnessing}. Note that \cite{riloff} do not report performance on this dataset and are hence, not included in this table. Figure~\ref{fig:optimal} shows how optimal threshold is determined. Note that similar trends are observed for other cases as well. In this case, our approaches do not perform as well as the past reported value in \cite{joshi2015harnessing}. Also, unlike the tweets, the Incongruous words-only approach results in a degradation as compared to all-words approach, for discussion forum posts. This shows that while our approach works for short text (tweets), it does not work for long text (discussion forum posts). This is because the average length of discussion forum posts is higher than that of tweets. As a result, the all-words approach or even Incongruous words-only approach may introduce similarity comparison with irrelevant words (`\textit{man}' and `\textit{woman}' in the example in Section 3). 


\begin{table}[]
\centering

\begin{tabular}{p{2.8cm}p{0.5cm}lll}
\toprule
 & &	\textbf{P}&	\textbf{R}&\textbf{F}\\ \midrule
Joshi et al. (2015) & & 48.9 & 92.4 & 64 \\
     \midrule
      \textbf{Similarity} 	&\textbf{T}&	\textbf{P}&	\textbf{R}&\textbf{F}\\ \midrule
\multicolumn{5}{c}{\textbf{All-Words Approach}}\\\midrule
Word2Vec	&0.48&	56.14&	52.17	&54.08\\
WordNet	&0.27&	45.12&	47.68&	46.37\\ \midrule
\multicolumn{5}{c}{\textbf{Incongruous words-only Approach}}\\\midrule
Word2Vec	&0.36	&37.04	&47.48&	41.61\\
WordNet&	0.15	&42.69&	48.18	&45.27	\\
\bottomrule
\end{tabular}
\caption{Results of our approach for dataset of discussion forum posts by Walker et al (2012), compared with best reported value on the same dataset}
\label{tab:res2}
\end{table}
\begin{table}[]
\centering
\begin{tabular}{p{1.6cm}cccc}
\toprule & & \textbf{P} & \textbf{R} & \textbf{F} \\ \midrule
\multicolumn{2}{c}{Riloff et al. (2013)} & 62 & 44 & 51 \\ 
\multicolumn{2}{c}{Joshi et al. (2015)} & 77 & 51 & 61 \\
\midrule
 \textbf{Similarity Metric}   & \textbf{Best-T} &  \textbf{P} & \textbf{R} & \textbf{F} \\ \midrule
\multicolumn{5}{c}{\textbf{All words}} \\ \midrule
\multirow{1}{*}{Word2Vec} & (0.1, 0.1)                        & 67.68       & 47.96       & 56.12       \\
\multirow{1}{*}{WordNet}     &  (0.1, 0.1)                      & 68.83       & 76.93       & 72.66       \\      \midrule
 \multicolumn{5}{c}{\textbf{Incongruous words-only}} \\ \midrule
\multirow{1}{*}{Word2Vec} &(0.42,0.1) &63.92 & 77.64& 70.09    \\
\multirow{1}{*}{WordNet}     &(0.14,0.12) &82.81 &77.91 & \textbf{80.28}       \\      \bottomrule
\end{tabular}
\caption{Two-fold cross-validation performance of our approaches for the tweets dataset; Best-T values in parentheses are optimal thresholds as obtained for the two folds}
\label{tab:twofallword}
\end{table}

                                   
\begin{table}[]
\centering
\begin{tabular}{p{1.6cm}cccc}
\toprule
  & & \textbf{P} & \textbf{R} & \textbf{F} \\ \midrule
\multicolumn{2}{c}{Joshi et al. (2015)} & 48.9 & 92.4 & \textbf{64} \\
      \midrule
\textbf{Similarity Metric}   & \textbf{Best-T} &  \textbf{P} & \textbf{R} & \textbf{F} \\ \midrule
 \multicolumn{5}{c}{\textbf{All words}} \\ \midrule
 \multirow{1}{*}{Word2Vec} &  (0.48, 0.48)                       & 56.20      & 52.17       & 54.10       \\
\multirow{1}{*}{WordNet}     & (0.37, 0.46)                       & 43.13       & 48.04       & 45.45      \\\midrule                        
  \multicolumn{5}{c}{\textbf{Incongruous words-only}} \\ \midrule
\multirow{1}{*}{Word2Vec} &(0.19,0.25) &36.48 &47.41 &41.23      \\
\multirow{1}{*}{WordNet}   &(0.15,0.12) &28.34 &48.04
&35.33 \\\bottomrule                              
\end{tabular}
\caption{Two-fold cross-validation performance of our approaches for the discussion forum posts dataset; Best-T values in parentheses are optimal thresholds as obtained for the two folds}
\label{tab:twoftopincr}
\end{table}
\subsection{Two-fold cross-validation}
Tables~\ref{tab:twofallword} and~\ref{tab:twoftopincr} show the two-fold cross-validation performance in case of tweets and discussion forum posts respectively. In each of the cases, past work that reports results on the same dataset is also mentioned: \cite{riloff} and \cite{joshi2015harnessing} report performance on the tweets dataset while \cite{joshi2015harnessing} do so on the discussion forums dataset. The optimal values of threshold for the two folds are also reported since they cannot be averaged. Table~\ref{tab:twofallword} shows that the incongruous words-only approach outperforms past work and the all words approach. The best performance is 80.28\% when incongruous words-only approach and WordNet similarity are used. Thus, in the case of tweets, our approaches perform better than past reported values.

Table~\ref{tab:twoftopincr} shows the corresponding values for the discussion forum posts. Unlike tweets, both our approaches do not perform as well as past reported values. The reported value of F-score is 64\% while our approaches achieve a best F-score of 54.10\%. This is likely because discussion forum posts are longer than tweets and hence, the set of candidate incongruous words is larger. This negative observation, in combination with the observation in case of tweets above, is an indicator of how the set of candidate incongruous words is a crucial parameter of the success of our approaches.

\begin{table}[]
\centering
\begin{tabular}{p{2.5cm}llll}
\toprule
\textbf{Approach} 	&\textbf{T}&	\textbf{P}&	\textbf{R}&	\textbf{F} \\ \midrule
All-words & 0.29 & 55.07 & 55.78 & 55.43\\
Oracle  & 0.014 & 59.13 & 68.37 & \textbf{63.42}\\
\bottomrule
\end{tabular}
\caption{Performance of the all-words approach versus the situation when the exact incongruous word is known}
\label{tab:res3}
\end{table}
\section{Discussion}
\label{sec:disc}
Since our approaches perform well for short text like tweets but not for long text such as discussion forum posts, choosing the right set of candidate positions appears to be crucial for the success of the proposed technique. The Incongruous words-only is a step in that direction, but we observe that it is not sufficient in case of discussion forum posts. Hence, in this section, we consider an oracle case: the exact incongruous word case. This is the case where the exact incongruous word is known. Hence, we now compare our all-words approach with an `\textit{exact incongruous word}' approach, when the exact incongruous word is known. In this case, we do not iterate over all word positions but only the position of the incongruous word. For the purpose of these experiments, we use the dataset by~\cite{ghoshsarcastic}. Their dataset consists of a word, a tweet containing the word and the sarcastic/non-sarcastic label. In case of sarcastic tweets, the word indicates the specific incongruous word. Table~\ref{tab:res3} compares the all-words approach with the only incongruous word approach. We observe that the F-score increases from 55.43\% to 63.42\% when the exact incongruous word is known. This shows that our approaches can be refined further to be able to zone in on a smaller set of candidate incongruous words.

It is never possible to know the exact incongruous word in a sentence. Therefore, future approaches that follow this line of work would need to work towards reducing the set of candidate incongruous words.

\section{Error Analysis}
\label{sec:erranal}
Some errors made by our approaches are due to the following reasons:
\begin{enumerate}
\item \textbf{Absence of WordNet senses}: For a certain input sentence, the word `\textit{cottoned}' is returned as the most likely word for a position. However, no sense corresponding to the word exists in WordNet, and so the word is ignored.
\item \textbf{Errors in sentence completion}: The sarcastic sentence `\textit{Thank you for the input, I'll take it to heart}\footnote{This tweet is labeled as sarcastic in the dataset by \cite{riloff}}' is incorrectly predicted as non-sarcastic. For the position where the word `\textit{input}' is present, the expected word as returned by context2vec is `\textit{message}'.
\end{enumerate}
 
\section{Conclusion \& Future Work}
\label{sec:concl}
This paper describes how sentence completion can be used for sarcasm detection. Using context2vec, a sentence completion toolkit, we obtain the expected word at a given position in a sentence, and compute the similarity between the observed word at that position and the expected word. Since the position of the incongruous (observed) word may not be known, we consider two approaches: (a) All-words approach in which context2vec is invoked for all content words, (b) Incongruous words-only approach where context2vec is invoked only for 50\% most incongruous words. We present our experiments on two datasets: tweets and book snippets, and for two similarity measures: word2vec similarity, and WordNet similarity. Our approach outperforms past reported work for tweets but not for discussion forum posts, demonstrating that sentence completion can be used for sarcasm detection of short text. Finally, we validate the benefit of our approach for an oracle case where the exact incongruous word is known. Our approach results in a 8\% higher F-score as compared to the all-words approach. Our error analysis shows that absent WordNet senses and errors in sentence completion results in errors by our approach.

Our findings set up the promise of sentence completion for sarcasm detection. This work can be extended by incorporating the current technique as a set of features for a statistical classifier. Since our approaches do not perform well for discussion forum posts, our approach must be refined to arrive at a good subset of candidate incongruous words.
\bibliographystyle{IEEEtran}
\bibliography{refs}
\end{document}